%% file: main.tex
\documentclass[11pt]{article}

\usepackage{acl}
\usepackage{float}
\usepackage{multirow}
\usepackage{times}
\usepackage{latexsym}
\usepackage{booktabs}
\usepackage[T1]{fontenc}
\usepackage{array, tabularx}
\usepackage[utf8]{inputenc}
\usepackage{tikz}
\usepackage{microtype}
\usepackage{tcolorbox}
\usepackage{fancyvrb} 
\usepackage{booktabs}
\usepackage{multirow}
\usepackage{colortbl}
\usepackage{xcolor}
\usepackage{listings}
\definecolor{dsbg}{HTML}{EAF1F8}            
\definecolor{dsoverall}{HTML}{CBDDF2}        

\definecolor{llamabg}{HTML}{EBF5EB}         
\definecolor{llamaoverall}{HTML}{CEEBCE}    

\definecolor{qw27bg}{HTML}{FAF4E4}          
\definecolor{qw27overall}{HTML}{F3E7C4}     

\definecolor{qw9bg}{HTML}{F8EAEF}           
\definecolor{qw9overall}{HTML}{F1D2DB}      

\definecolor{qw4bg}{HTML}{ECE7FA}           
\definecolor{qw4overall}{HTML}{D9CEF4}      

\definecolor{avgback}{HTML}{F5F5F5}         
\definecolor{avgover}{HTML}{E5E5E5}         

\usepackage{graphicx}
\usepackage{amsmath}

\title{SchemaGUI: A Schema-Driven Benchmark for Controllable GUI Generation Evaluation}

\author{
  \textbf{Jiarui Dong\textsuperscript{1}},
  \textbf{Yin Cai\textsuperscript{2}},
  \textbf{Zhouhong Gu\textsuperscript{2}},
  \textbf{Chenmou Wu\textsuperscript{1}},
  \textbf{Ci Tao\textsuperscript{2}},
  \\
  \textbf{Yiran Chen\textsuperscript{2}},
  \textbf{Jialing Li\textsuperscript{2}},
  \textbf{Xiaoran Shi\textsuperscript{2}},
  \textbf{Juntao Zhang\textsuperscript{2}},
  \textbf{Zhijun Fang\textsuperscript{2,*}}
  \\[3pt]
  \textsuperscript{1}Shanghai University of Engineering Science
  \quad
  \textsuperscript{2}Fudan University
  \\[3pt]
  {\small
  \{M320124105, chenmou0410\}@sues.edu.cn}
  \\
  {\small
  \{ycai25, zhgu22, ctao24, 25213050128\}@m.fudan.edu.cn}
  \\
  {\small
  \{jialingli22, xrshi25, 26213050478\}@m.fudan.edu.cn}
  \\[2pt]
  {\small
  \textsuperscript{*}\textbf{Corresponding author:}
  zjfang@fudan.edu.cn}
}

\begin{document}


\maketitle
\begin{abstract}
\input{latex/00.abstract.03}
\end{abstract}

\section{Introduction}
\input{latex/01.Introduction.06}

\section{Related Work}
\input{latex/02_related_work}

\section{Methodology}
\input{latex/03.Methodology.05}

\section{Experiment Setup}
\input{latex/04_experiment_setup}

\section{Analysis}
\input{latex/05.Analysis.05}

\section{Conclusion}
\input{latex/06.Conclusion.01}

\section*{Limitations}
\input{latex/07.Limitation.02}

\section*{Ethical Concern}
\input{latex/09_ethical_concern}

%

\bibliography{custom}

\appendix

\section{Appendix}
\label{sec:appendix}
\input{latex/08.Appendix.02}

\end{document}

%% file: latex/00.abstract.03.tex
Large language models (LLMs) have demonstrated strong potential in graphical user interface (GUI) generation, but reliable evaluation remains challenging due to uncontrolled data distributions, noisy annotations, and limited layout scenario coverage. To address this, we propose SchemaGUI, a template-based benchmark for controllable GUI generation evaluation. By synthesizing paired natural language instructions and deterministic function-call references from parameterized interface schemas, SchemaGUI can generate thousands of deterministically annotated tasks in seconds without human labeling. Based on 1,000 evaluated instances per scenario and language across six representative bilingual scenarios, we benchmark five mainstream models, including the Qwen3.5 family, Qwen3-Coder-30B, and DeepSeek-R1. Our extensive analysis reveals three key insights. First, precise geometric spatial control remains an important bottleneck; while scaling Qwen3.5 from 4B to 27B improves Schema Feasibility from 91.56\% to 99.63\%, the Geometry score improves more modestly (from 67.05\% to 75.30\%). Second, generation difficulty is highly sensitive to layout complexity, with current LLMs excelling at simple sequential arrangements but suffering severe coordinate drift in dense grids and multi-region compositions. Third, thinking mode increases token consumption while generally reducing GUI Score, particularly for smaller models. Our code is available at
\href{https://github.com/xdong2002/SchemaGUI}{github.com/xdong2002/SchemaGUI}.

%% file: latex/01.Introduction.06.tex
Large Language Models (LLMs) have achieved substantial progress in Graphical User Interface (GUI) generation. Early frameworks, such as pix2code~\citep{beltramelli2018pix2code} and Sketch2Code~\citep{robinson2019sketch2codegeneratingwebsitepaper}, demonstrated the feasibility of end-to-end generation by translating screenshots or hand-drawn sketches into code. With the advent of modern multimodal models, approaches like Design2Code~\citep{si2024design2code} and LayoutGPT~\citep{feng2023layoutgptcompositionalvisualplanning} have pushed the generation targets toward high-fidelity web pages and layouts, while WAFFLE~\citep{liang2026wafflefinetuningmultimodalmodels}, ScreenCoder~\citep{jiang2025screencoderadvancingvisualtocodegeneration}, and UICopilot~\citep{gui2025uicopilot} have advanced structural alignment, modular reasoning, and functional synthesis. Consequently, generated outputs have evolved from visually similar mockups to fully structured, interactive interfaces that can be rendered in a browser.

\begin{figure*}[t]
\centering
\includegraphics[width=0.98\linewidth]{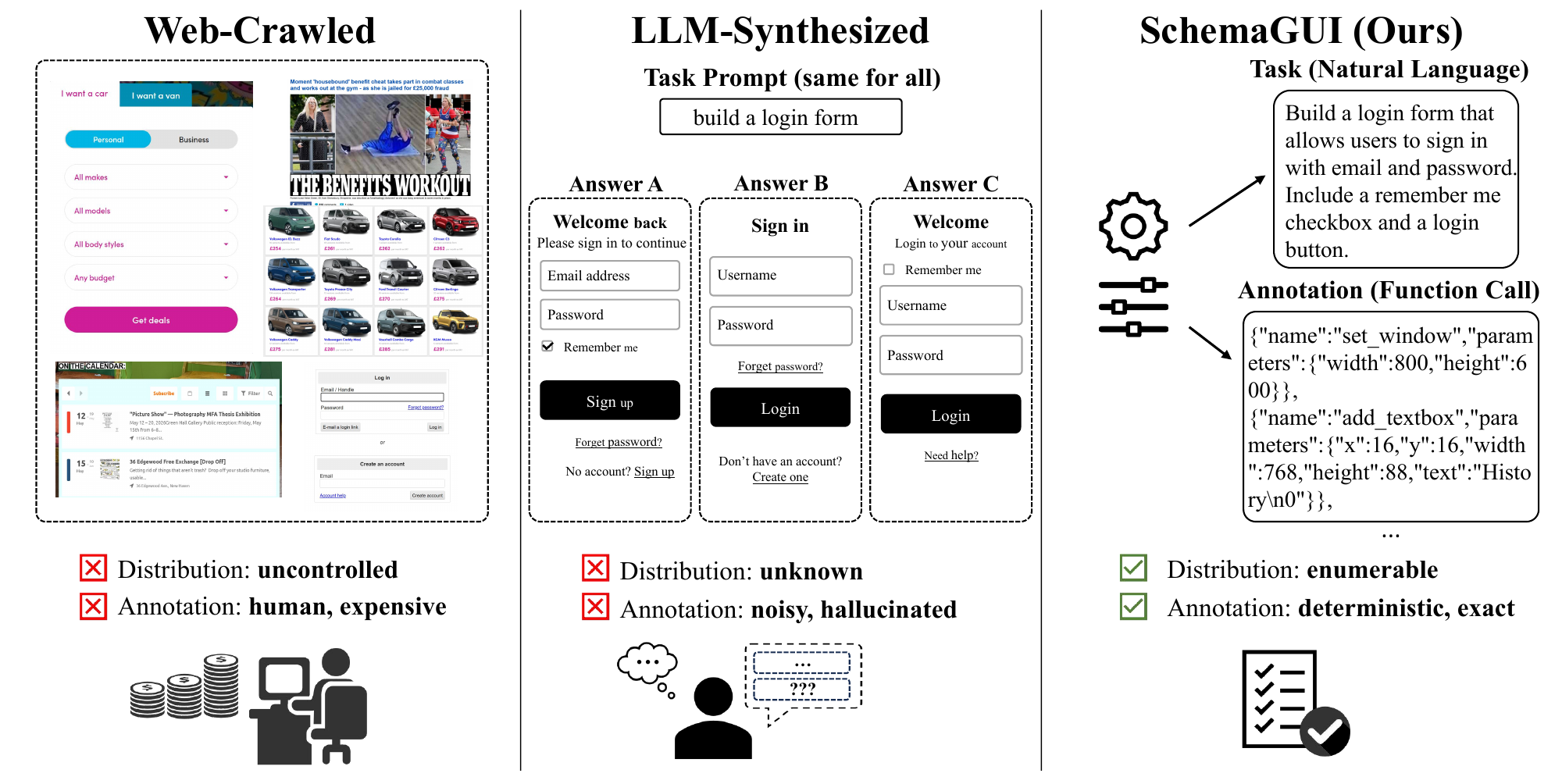} 
\caption{Comparison of different data construction paradigms for GUI generation evaluation. (Left) \textbf{Web-Crawled}: data distribution is uncontrolled with expensive manual annotations. (Middle) \textbf{LLM-Synthesized}: the data distribution remains unknown while outputs suffer from model noise and hallucinated references. (Right) \textbf{SchemaGUI (Ours)}: parametric slot configurations enable enumerable distributions and deterministic, exact annotations.}
\label{fig:teaser}
\end{figure*}

Evaluating the GUI generation model requires a benchmark dataset that satisfies two fundamental criteria: comprehensive scenario coverage and clean individual samples. As illustrated in Figure~\ref{fig:teaser}, current benchmarks fall short on both counts. One approach relies on web crawling and manual annotation, which produces uncontrolled samples, limited scenario diversity, and high annotation costs, making large-scale evaluation uncontrolled. Furthermore, the high manual annotation costs make scaling this approach unsustainable~\citep{gu2024structextevalevaluatinglargelanguage}. However, these synthetic samples inherit model hallucinations and biases, leading to unreliable evaluation results. Another approach uses LLMs to generate data at scale, but the synthetic samples inherit model noise, hallucinations, and biases, making the evaluation targets inherently unclean. This unstable data quality and coverage ultimately make the model comparisons unreliable. What GUI evaluation lacks is a dataset that is coverage-controlled, clean per sample, and scalable. Figure~\ref{fig:teaser} contrasts existing data-construction paradigms with SchemaGUI.

To address this, we propose SchemaGUI, a template-based framework for automatically generating GUI evaluation data. SchemaGUI covers six common interface types, including calculator, login form, music player, document editor, image editor, and information card, with adjustable layouts, components, and text. Each configuration produces two paired outputs: a natural language task description and a component-wise annotation listing positions, sizes, and text. Using this method, we generated a dataset of 1,000 tasks with bilingual (Chinese-English) instructions and cross-backend validation. Models are asked to produce outputs in the same structured format, allowing line-by-line comparison with the reference. The process can generate more than 10,000 samples in seconds, with all annotations derived directly from the template, without relying on human labeling or model outputs.

Based on this dataset, comprising 6,000 evaluation instances per language and 12,000 bilingual instances for each evaluated model, we conducted a systematic evaluation of five mainstream large language models: Qwen3.5-4B, Qwen3.5-9B, Qwen3.5-27B, Qwen3-Coder-30B, and DeepSeek-R1. The experimental results reveal three key insights. First, \textbf{precise geometric spatial control remains an important challenge for current models}. As the model size scales from Qwen3.5-4B to 27B, its average Schema Feasibility increases from 91.56\% to 99.63\%, while its average Geometry score improves more modestly from 67.05\% to 75.30\%. Second, \textbf{generation difficulty is highly sensitive to layout complexity}. Existing models perform better on simple linear layouts but experience substantial performance degradation in scenarios involving dense components (e.g., calculators) or multi-region combinations. Finally, \textbf{thinking mode negatively affects GUI generation performance in the evaluated Qwen3.5 models}. Enabling thinking mode for Qwen3.5-27B reduces the Overall GUI Score by 9.72 percentage points in EN and 12.36 percentage points in ZH while substantially increasing response-token consumption.

%% file: latex/02_related_work.tex
\subsection{Text-to-UI Generation and End-to-End Models}

The development of modern GUI generation techniques originates from early neural layout reconstruction methods. Representative works include pix2code~\citep{beltramelli2018pix2code}, which transforms GUI screenshots into structured code, and Sketch2Code~\citep{robinson2019sketch2codegeneratingwebsitepaper}, which further enables the automatic generation of HTML from hand-drawn paper UI sketches.

Building upon these foundations, high-fidelity synthesis has emerged via MLLMs. For instance, Generative UI~\citep{leviathan2025generative} highlights that LLMs are effective zero-shot UI generators, and Design2Code~\citep{si2024design2code} establishes a comprehensive framework for evaluating the translation of visual designs into production-ready code. Beyond direct mapping, recent research focuses on structural alignment and task flexibility. WAFFLE~\citep{liang2026wafflefinetuningmultimodalmodels} introduces a structure-aware attention mechanism to better align UI images with HTML hierarchical logic. Meanwhile, UI-UG~\citep{yang2025uiugunifiedmllmui} unifies diverse tasks, including image-to-UI and text-to-code, under a single instruction-following framework. PSD2Code~\citep{chen2025psd2codeautomatedfrontendcode} explores direct functional mapping from structured design files to front-end logic.

However, these end-to-end paradigms rely on stochastic generation without explicit structural constraints, often causing layout drift or component overlap. To ensure geometric determinism, we introduce a schema-constrained functional intermediate representation (IR).

\subsection{Intermediate Representations and Structured Specification}

To enhance controllability, researchers have explored structural modeling of UI hierarchies. Early probabilistic frameworks like LayoutVAE~\citep{jyothi2021layoutvaestochasticscenelayout} propose VAE-based architectures for stochastic scene layout generation, capturing both object counts and spatial relationships within UI structures. Similarly, HSSN~\citep{li2022deep} utilizes multi-label hierarchy constraints and margin-induced embedding refinement to enforce tree-structured dependencies.

To further constrain the generative process, UI grammar~\citep{lu2023uilayoutgenerationllms} guides LLMs in maintaining rigorous syntactic consistency during layout synthesis. To rectify latent layout deficiencies, AlignUI~\citep{liu2026alignuimethoddesigningllmgenerated} employs preference-constrained modeling and structural alignment strategies to enhance generative stability. Regarding design abstractions, SpecifyUI~\citep{chen2024specifyui} introduces structured specifications to support the iterative expression of design intent. Recent critique-based frameworks like UICrit~\citep{duan2024uicrit} further emphasize the importance of structural feedback.

However, current models often mimic design patterns without understanding physical layout rules, leading to geometric conflicts like component overlap. We address this by incorporating a geometric audit on predicted function calls to ensure physical feasibility.

\subsection{Programmatic Generation and Functional Synthesis}

To move beyond static structural modeling, recent advancements focus on the execution-oriented logic of interfaces. This shift was catalyzed by benchmarks like META-GUI~\citep{sun2022metaguimultimodalconversationalagents}, which frame UI tasks as multimodal programmatic action sequences across conversational interactions. WebArena~\citep{zhou2024webarena} establishes a self-hosting web environment to evaluate the ability of agents to execute complex, multi-step programmatic workflows.

To handle the escalating complexity of these tasks, recent frameworks have introduced modular reasoning strategies. GUIDE~\citep{kolthoff2025guidellmdrivenguigeneration} addresses reliability through task decomposition, breaking down high-level UI requirements into manageable sub-goals. CANVAS~\citep{jeong2025canvasbenchmarkvisionlanguagemodels} evaluates models within a tool-based design paradigm, focusing on the iterative use of external software to refine design intent. These advancements have fostered emerging synthesis frameworks, where UICopilot~\citep{gui2025uicopilot} leverages context-aware code synthesis to bridge the gap between design and development, and Widget2Code~\citep{zhang2025widget2code} utilizes component-level function calling to map visual widgets directly into executable code.

Nevertheless, these atomized methods often focus on localized tasks rather than maintaining the systemic integrity of the entire layout. We bridge this gap by proposing a unified functional synthesis paradigm that governs the full layout through formal parameter space modeling.

\begin{figure*}[t]
\centering
\includegraphics[width=\textwidth]{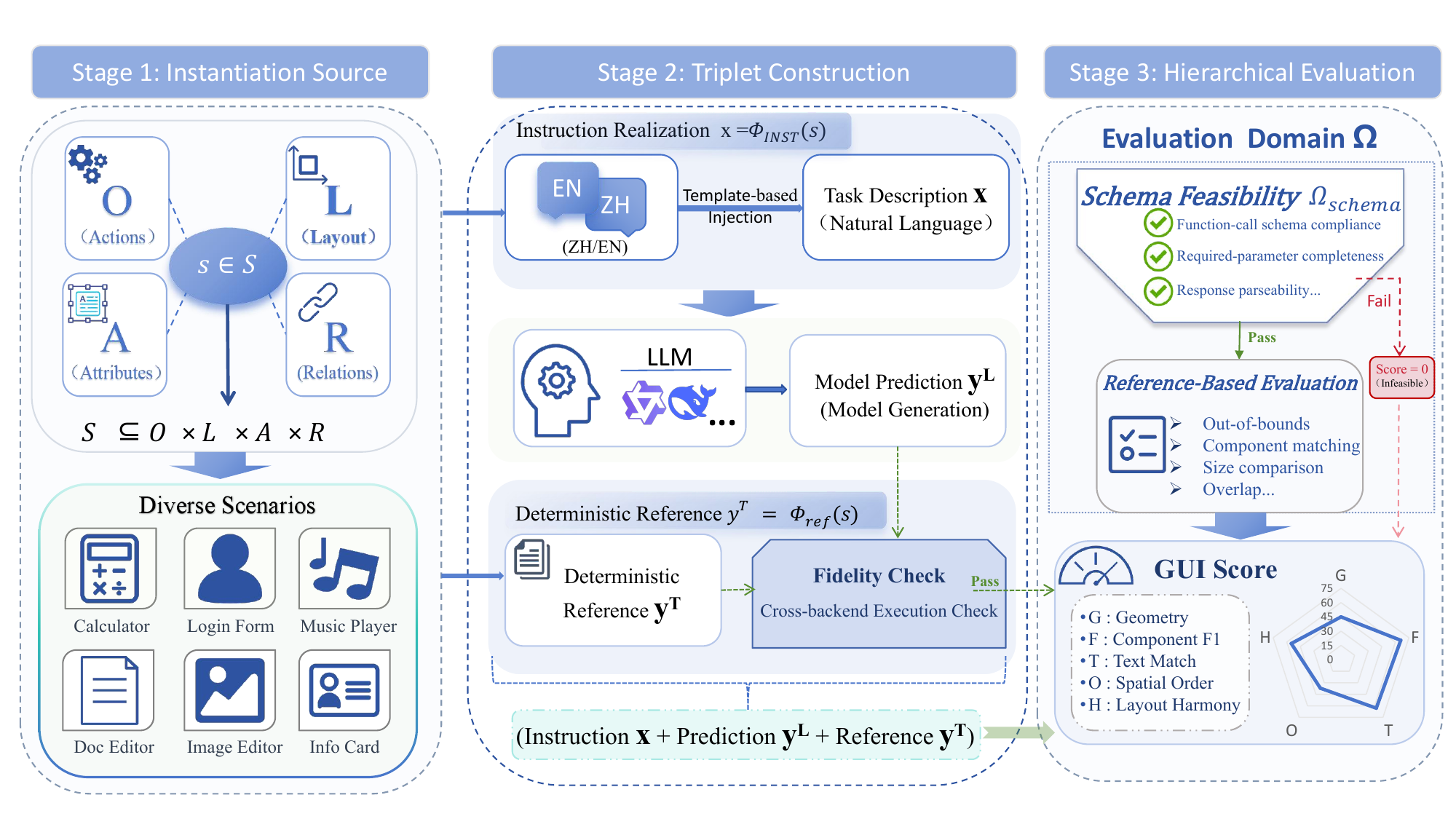}
\caption{Overview of the SchemaGUI benchmark. Stage 1 samples a GUI instance \(s\) from the schema-driven design space \(\mathcal{S}\). Stage 2 constructs the aligned triplet \((x,y^T,y^L)\) through instruction realization, deterministic reference generation, and model prediction. Stage 3 assigns zero to schema-infeasible predictions and computes the GUI Score for feasible predictions over five evaluation dimensions.}
\label{fig:framework}
\end{figure*}

%% file: latex/03.Methodology.05.tex
\subsection{Formalization of the Schema-Driven GUI Design Space}
\label{sec:formalization}

To address the low efficiency of acquiring high-quality data, we formalize the GUI generation task within a controllable and strictly standardized framework.

In our framework, we conceptualize a Graphical User Interface (GUI) as a structured collection of components, where the generation process is defined as a sequence of atomic function calls. Specifically, we introduce a functional Intermediate Representation (IR) defined as a sequence of atomic function calls \(y\):
\begin{equation}
y=[c_1,c_2,\dots,c_n].
\label{eq:component_sequence}
\end{equation}
Each individual call \(c_i=(f_i,\theta_i)\) serves as an independent generative instruction. The operation primitive \(f_i\) specifies the component type, such as \texttt{add\_button} or \texttt{add\_textbox}, while the associated parameters \(\theta_i\) provide the specific details. These parameters consist of geometric attributes from \(\mathcal{L}\), defining component coordinates and dimensions, and content attributes from \(\mathcal{A}\), covering text labels and data values.

Each individual call \(c_i\) represents an atomic operation. However, simply stacking valid calls does not automatically result in a functional GUI, since the complete interface also depends on the ordering and spatial relations among components. To control the generation of these sequences \(y\), we introduce the schema-driven design space:
\begin{equation}
\mathcal{S}\subseteq\mathcal{O}\times\mathcal{L}\times\mathcal{A}\times\mathcal{R}.
\label{eq:schema_space}
\end{equation}
The design space \(\mathcal{S}\) is constructed by exhaustively enumerating compatible parameter combinations, with each valid combination defining a design instance \(s\in\mathcal{S}\). Formally, each instance \(s\) is represented as a combination of four functional domains:

\textbf{Operational Actions} \((\mathcal{O})\) represent the actions of the construction process, covering environment initialization and component instantiation. \textbf{Layout Geometry} \((\mathcal{L})\) specifies spatial placement and sizing, defining precise coordinates and structural patterns such as grids. \textbf{Content Attributes} \((\mathcal{A})\) provide the actual content for the components, filling slots with semantic data such as button labels or display text. \textbf{Structural Relations} \((\mathcal{R})\) specify the rules of organization, governing component grouping and alignment to ensure structural consistency.

Building upon the formalized instance \(s\), the next stage of our framework translates it into two distinct forms: a model-readable input and a verifiable reference. The model prediction is subsequently generated from the model-readable input:
\begin{align}
x &= \Phi_{\mathrm{inst}}(s), \label{eq:inst_generation}\\
y^T &= \Phi_{\mathrm{ref}}(s), \label{eq:ref_generation}\\
y^L &= \Psi_{\mathrm{LLM}}(x). \label{eq:llm_generation}
\end{align}

\noindent\textbf{Instruction Realization} \((s\mapsto x)\): The instance \(s\) is materialized into a natural-language task description \(x\) through template-based injection, serving as the input to the evaluated model.

\noindent\textbf{Reference Realization} \((s\mapsto y^T)\): Simultaneously, the same instance \(s\) is deterministically expanded into a reference function-call sequence \(y^T\) following the scenario-specific schema.

Given the instruction \(x\), the model generates a predicted sequence \(y^L\). To ensure a fair assessment, we associate the prediction with the instruction and its deterministic reference to form the evaluation triplet \((x,y^T,y^L)\).

\subsection{Schema-Conditioned GUI Evaluation}
\label{sec:schema_conditioned_eval}

Given the design space defined in Section~\ref{sec:formalization}, we evaluate the generated GUI artifacts as follows.

We define \(\Omega\) as the index set of all aligned evaluation samples, where each \(i\in\Omega\) corresponds to a triplet \((x_i,y_i^T,y_i^L)\). Here, \(x_i\) denotes the input of the \(i\)-th sample, \(y_i^T\) denotes its deterministic reference, and \(y_i^L\) denotes the corresponding model prediction. We further define \(\Omega_{\mathrm{schema}}\subseteq\Omega\) as the schema-feasible domain, consisting of sample indices whose predictions can be parsed into structurally valid and executable function-call sequences.

\textbf{Schema feasibility} captures whether the model can produce structurally executable function-call sequences. Samples with empty responses, request failures, unrecoverable parsing errors, invalid call-sequence formats, or missing required arguments fall outside \(\Omega_{\mathrm{schema}}\) and are assigned a GUI Score of zero.

For samples indexed by \(\Omega_{\mathrm{schema}}\), each prediction \(y_i^L\) is compared with its deterministic reference \(y_i^T\) using five submetrics: Geometry \((\mathcal{G}_i)\), Component F1 \((\mathcal{F}_i)\), Text Match \((\mathcal{T}_i)\), Spatial Order \((\mathcal{O}_i)\), and Layout Harmony \((\mathcal{H}_i)\).

\textbf{Geometry} \((\mathcal{G}_i)\): geometric fidelity, combining coordinate and size consistency with penalties for unmatched reference components, out-of-bounds placement, and severe overlap. \textbf{Component F1} \((\mathcal{F}_i)\): set-level completeness and correctness, measured by comparing the types and quantities of predicted and reference components. \textbf{Text Match} \((\mathcal{T}_i)\): exact consistency of normalized \texttt{text} and \texttt{data} values within each component type. \textbf{Spatial Order} \((\mathcal{O}_i)\): preservation of the relative top-to-bottom and left-to-right order among matched components. \textbf{Layout Harmony} \((\mathcal{H}_i)\): visual harmony of the generated layout in terms of component alignment, inter-component spacing, and relative size proportions.

Detailed component matching and metric implementation are provided in Appendix~\ref{app:metric_details}. The decomposition of Geometry and its layout penalties is presented in Appendix~\ref{app:geometry_score}.

The final GUI Score \((\mathrm{GS})\) is defined as
\begin{equation}
\mathrm{GS}_i=
\begin{cases}
\displaystyle\sum_{k=1}^{5}w_k m_{i,k}, & i\in\Omega_{\mathrm{schema}},\\[4pt]
0, & i\notin\Omega_{\mathrm{schema}},
\end{cases}
\label{eq:gui_score}
\end{equation}
where \(\mathbf{m}_i=[\mathcal{G}_i,\mathcal{F}_i,\mathcal{T}_i,\mathcal{O}_i,\mathcal{H}_i]\) is the submetric vector for the \(i\)-th sample, \(m_{i,k}\) denotes its \(k\)-th submetric, and \(w_k\) is the corresponding weight. We use equal weights, setting \(w_k=1/5\) for \(k\in\{1,\ldots,5\}\). Thus, the GUI Score for a schema-feasible prediction is the mean of the five submetrics.This evaluation protocol focuses on structured GUI generation under explicit schema and layout constraints rather than open-ended interface design.

\definecolor{coderbg}{RGB}{232,241,234}
\definecolor{coderoverall}{RGB}{214,230,218}

\begin{table*}[t]
\centering
\small
\setlength{\tabcolsep}{3.6pt}
\renewcommand{\arraystretch}{1.25}

\begin{tabular}{l c ccccccc}
\toprule
\textbf{Model}
& \textbf{Lang.}
& \textbf{Calculator}
& \textbf{Login Form}
& \textbf{Music Player}
& \textbf{Doc Editor}
& \textbf{Image Editor}
& \textbf{Info Card}
& \textbf{Overall} \\
\midrule

\multirow{2}{*}{Qwen3.5-4B}
& EN
& \cellcolor{qw4bg}33.25
& \cellcolor{qw4bg}84.58
& \cellcolor{qw4bg}77.84
& \cellcolor{qw4bg}87.34
& \cellcolor{qw4bg}74.97
& \cellcolor{qw4bg}82.59
& \cellcolor{qw4overall}73.43 \\

& ZH
& \cellcolor{qw4bg}56.26
& \cellcolor{qw4bg}89.92
& \cellcolor{qw4bg}84.20
& \cellcolor{qw4bg}78.79
& \cellcolor{qw4bg}65.35
& \cellcolor{qw4bg}\textbf{84.71}
& \cellcolor{qw4overall}76.54 \\
\midrule

\multirow{2}{*}{Qwen3.5-9B}
& EN
& \cellcolor{qw9bg}43.40
& \cellcolor{qw9bg}85.14
& \cellcolor{qw9bg}\underline{87.51}
& \cellcolor{qw9bg}\textbf{88.88}
& \cellcolor{qw9bg}64.26
& \cellcolor{qw9bg}83.77
& \cellcolor{qw9overall}75.49 \\

& ZH
& \cellcolor{qw9bg}63.22
& \cellcolor{qw9bg}90.25
& \cellcolor{qw9bg}\textbf{88.24}
& \cellcolor{qw9bg}\underline{79.53}
& \cellcolor{qw9bg}67.24
& \cellcolor{qw9bg}78.74
& \cellcolor{qw9overall}77.87 \\
\midrule

\multirow{2}{*}{Qwen3.5-27B}
& EN
& \cellcolor{qw27bg}\underline{71.93}
& \cellcolor{qw27bg}\textbf{93.07}
& \cellcolor{qw27bg}\textbf{88.49}
& \cellcolor{qw27bg}\underline{88.85}
& \cellcolor{qw27bg}\underline{80.84}
& \cellcolor{qw27bg}\underline{84.51}
& \cellcolor{qw27overall}\textbf{84.61} \\

& ZH
& \cellcolor{qw27bg}\underline{74.82}
& \cellcolor{qw27bg}\textbf{93.10}
& \cellcolor{qw27bg}86.03
& \cellcolor{qw27bg}\textbf{81.38}
& \cellcolor{qw27bg}\underline{70.81}
& \cellcolor{qw27bg}\underline{84.55}
& \cellcolor{qw27overall}\textbf{81.78} \\
\midrule

\multirow{2}{*}{Qwen3-Coder-30B}
& EN
& \cellcolor{coderbg}\textbf{74.75}
& \cellcolor{coderbg}\underline{92.00}
& \cellcolor{coderbg}86.39
& \cellcolor{coderbg}82.56
& \cellcolor{coderbg}75.85
& \cellcolor{coderbg}\textbf{84.75}
& \cellcolor{coderoverall}\underline{82.72} \\

& ZH
& \cellcolor{coderbg}\textbf{75.54}
& \cellcolor{coderbg}92.21
& \cellcolor{coderbg}86.13
& \cellcolor{coderbg}78.04
& \cellcolor{coderbg}68.31
& \cellcolor{coderbg}82.65
& \cellcolor{coderoverall}\underline{80.48} \\
\midrule

\multirow{2}{*}{DeepSeek-R1}
& EN
& \cellcolor{dsbg}63.88
& \cellcolor{dsbg}89.92
& \cellcolor{dsbg}85.60
& \cellcolor{dsbg}39.41
& \cellcolor{dsbg}\textbf{84.60}
& \cellcolor{dsbg}83.78
& \cellcolor{dsoverall}74.53 \\

& ZH
& \cellcolor{dsbg}64.28
& \cellcolor{dsbg}\underline{92.54}
& \cellcolor{dsbg}\underline{86.23}
& \cellcolor{dsbg}75.00
& \cellcolor{dsbg}\textbf{72.63}
& \cellcolor{dsbg}82.46
& \cellcolor{dsoverall}78.86 \\
\bottomrule
\end{tabular}

\caption{Main GUI Score results (\%) across six GUI generation scenarios. EN and ZH denote the English and Chinese evaluation sets, respectively. Overall is the macro-average over the six scenarios, with each scenario weighted equally. Within each language, the best result in each column is shown in \textbf{bold}, and the second-best is \underline{underlined}.}
\label{tab:main_gui_results}
\end{table*}

%% file: latex/04_experiment_setup.tex
\subsection{Baseline Models}
\label{sec:baseline_models}

We evaluate five LLMs across different model families and parameter scales. The evaluated models include \texttt{Qwen3.5-4B}, \texttt{Qwen3.5-9B}, and \texttt{Qwen3.5-27B} \citep{qwen3.5}, \texttt{Qwen3-Coder-30B-A3B-Instruct} \citep{yang2025qwen3technicalreport}, and \texttt{DeepSeek-R1} \citep{Guo_2025}.

For each evaluation sample, the deterministic template output \(y_i^T\) serves as the reference against which the model prediction \(y_i^L\) is evaluated. It is not treated as an additional model baseline.

\subsection{Test Scenarios and Dataset Stratification}
\label{sec:test_scenarios}

We evaluate schema-aligned GUI generation under two language settings: English (EN) and Chinese (ZH). The two subsets use the same scenario definitions, function-call schemas, and layout controls, while the instructions and language-dependent content are instantiated in the corresponding language. Results for EN and ZH are reported separately. 

The template parameter pools define more than \(10{,}000\) valid candidate combinations for each scenario. For LLM evaluation, we use a fixed random seed to sample instances from the complete design space and remove instances that produce duplicate function-call layouts. We retain \(1{,}000\) instances for each scenario and language setting, resulting in \(6{,}000\) instances per language and \(12{,}000\) bilingual instances for each evaluated model.

The benchmark covers six scenarios with different layout characteristics. \texttt{Calculator} focuses on dense grid alignment; \texttt{Login Form} and \texttt{Info Card} use relatively simple sequential layouts; \texttt{Music Player} requires coordination among multiple functional components; and \texttt{Doc Editor} and \texttt{Image Editor} contain multi-region layouts requiring more precise spatial organization. Each instance is generated from a parameterized template, providing deterministic alignment between the instruction \(x_i\) and its reference function-call sequence \(y_i^T\). Detailed scenario schemas, parameter dimensions, and generation rules are provided in Appendix~\ref{sec:scenario_schemas}.

\subsection{Evaluation Metrics}
\label{sec:evaluation_metrics}

We evaluate the models in terms of GUI quality and generation efficiency.

\noindent\textbf{GUI quality.} We use the GUI Score \((\mathrm{GS})\) and its five submetrics defined in Section~\ref{sec:schema_conditioned_eval}: Geometry, Component F1, Text Match, Spatial Order, and Layout Harmony. Schema-infeasible predictions receive a GUI Score of zero, while schema-feasible predictions are evaluated using the equally weighted mean of the five submetrics.

\noindent\textbf{Generation efficiency.} We record the token consumption of each sample, including Prompt Tokens \((T_{\mathrm{in}})\) and Response Tokens \((T_{\mathrm{out}})\). Prompt Tokens measure the length of the model input, while Response Tokens measure the length of the generated output.

%% file: latex/05.Analysis.05.tex
\setlength{\textfloatsep}{8pt plus 2pt minus 2pt}
\begin{figure}[t]
    \centering
    \includegraphics[width=\columnwidth]{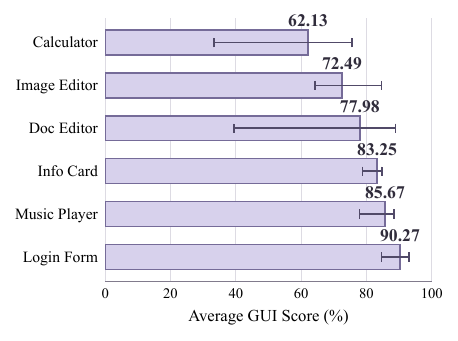}
    \caption{GUI Score across the six GUI generation scenarios, macro-averaged over the five evaluated models and two language settings. The horizontal lines indicate the minimum and maximum scores observed for each scenario.}
    \label{fig:scenario_score}
    \vspace{-5pt}
\end{figure}

\subsection{Overall Model Performance}
\label{sec:overall_performance}

\textbf{Qwen3.5-27B achieves the strongest overall performance in both language settings.} As shown in Table~\ref{tab:main_gui_results}, Qwen3.5-27B obtains the highest Overall GUI Score, reaching 84.61\% in EN and 81.78\% in ZH. Qwen3-Coder-30B ranks second in both settings, with Overall GUI Scores of 82.72\% and 80.48\%, respectively. The consistent ranking of these two models indicates that they achieve stronger overall performance under the structured function-call and layout requirements of SchemaGUI.

Within the Qwen3.5 family, the performance gain from model scaling is not uniform. From 4B to 9B, the Overall GUI Score increases from 73.43\% to 75.49\% in EN and from 76.54\% to 77.87\% in ZH. A larger improvement is observed from 9B to 27B, particularly in EN, where the score increases by 9.12 percentage points. DeepSeek-R1 achieves 74.53\% in EN and 78.86\% in ZH, remaining below Qwen3.5-27B and Qwen3-Coder-30B in overall performance.

\subsection{Performance Across GUI Scenarios}
\label{sec:scenario_performance}

\textbf{Simple sequential layouts are easier for current models, while dense grids and multi-region compositions present greater generation challenges.}Figure~\ref{fig:scenario_score} summarizes the average GUI Score for each scenario across the five evaluated models and two language settings.

As shown in Figure~\ref{fig:scenario_score}, \textit{Login Form} achieves the highest average GUI Score of 90.27\%, followed by \textit{Music Player} at 85.67\% and \textit{Info Card} at 83.25\%. These scenarios generally organize components through clear sequential arrangements or functionally separated groups, making their spatial relationships relatively straightforward to preserve.

In contrast, \textit{Calculator} obtains the lowest average score of 62.13\% and exhibits the widest performance variation, ranging from 33.25\% to 75.54\%. Its dense button grid requires consistent component sizes, spacing, and row--column alignment, so local coordinate deviations can disrupt the regularity of the complete layout. The multi-region \textit{Image Editor} and \textit{Doc Editor} scenarios also achieve lower average scores of 72.49\% and 77.98\%, respectively. These interfaces require models to coordinate multiple functionally distinct regions, such as toolbars, side panels, editing areas, and control groups. The particularly wide variation in \textit{Doc Editor} further suggests that models differ substantially in their ability to maintain such region-level organization.

\subsection{Performance Across Evaluation Dimensions}
\label{sec:performance across ui}

\begin{table}[h]
\centering
\footnotesize
\setlength{\tabcolsep}{2.2pt}
\renewcommand{\arraystretch}{1.12}
\begin{tabular*}{\columnwidth}{@{\extracolsep{\fill}}lrrrrrr}
\toprule
\textbf{Model} & \(\mathcal{G}\) & \(\mathcal{F}\) & \(\mathcal{T}\) & \(\mathcal{O}\) & \(\mathcal{H}\) & \textbf{Feas.} \\
\midrule
\multicolumn{7}{c}{\textbf{English (EN)}} \\
\midrule
Qwen3.5-4B    & 68.52 & 95.55 & 86.34 & \underline{84.38} & 70.10 & 86.07 \\
Qwen3.5-9B    & 73.33 & 94.68 & 86.15 & 82.07 & 70.40 & 88.68 \\
Qwen3.5-27B   & \underline{77.37} & \underline{99.57} & \textbf{88.19} & \textbf{85.34} & \textbf{74.97} & \underline{99.37} \\
Qwen3-Coder   & 72.98 & \textbf{99.99} & \underline{87.08} & 82.20 & 72.12 & \textbf{99.83} \\
DeepSeek-R1   & \textbf{79.09} & 99.20 & 87.01 & 84.03 & \underline{74.28} & 87.65 \\
\midrule
\multicolumn{7}{c}{\textbf{Chinese (ZH)}} \\
\midrule
Qwen3.5-4B    & 65.57 & 94.43 & 85.59 & \textbf{79.75} & 67.50 & 97.05 \\
Qwen3.5-9B    & 70.98 & 96.10 & 81.32 & 77.31 & 69.56 & 98.32 \\
Qwen3.5-27B   & \underline{73.22} & 96.90 & \textbf{87.46} & \underline{79.36} & \textbf{72.34} & \textbf{99.90} \\
Qwen3-Coder   & 71.11 & \textbf{97.39} & \underline{87.44} & 76.28 & \underline{71.04} & \underline{99.82} \\
DeepSeek-R1   & \textbf{73.93} & \underline{97.19} & 87.19 & 75.56 & 70.73 & 97.25 \\
\bottomrule
\end{tabular*}
\caption{Submetric performance (\%) and Schema Feasibility (Feas.) in EN and ZH. Submetrics are averaged over schema-feasible predictions. Best and second-best results within each language are shown in \textbf{bold} and \underline{underlined}, respectively.}
\label{tab:submetric_analysis}
\end{table}

\textbf{Current LLMs still struggle with precise geometric layout control.} As shown in Table~\ref{tab:submetric_analysis}, Component F1 remains consistently high across models and language settings, whereas Geometry and Layout Harmony generally receive lower scores. This contrast indicates that, among schema-feasible predictions, the evaluated models can usually generate the required component types and quantities, but remain less reliable in controlling component positions, alignment, spacing, and relative size.

Qwen3.5-27B achieves the strongest overall balance across the five evaluation dimensions, together with near-perfect Schema Feasibility in both EN and ZH. Qwen3-Coder also achieves high Schema Feasibility and Component F1, but obtains lower Geometry and Layout Harmony scores. DeepSeek-R1 achieves the highest Geometry score in both language settings, although its lower Schema Feasibility in EN limits its Overall GUI Score. These results suggest that geometric layout control remains an important challenge even when models successfully produce schema-feasible function-call sequences.

\definecolor{coderbg}{RGB}{232,241,234}
\definecolor{coderoverall}{RGB}{214,230,218}

\begin{table*}[t]
\centering
\small
\renewcommand{\arraystretch}{1.12}
\setlength{\tabcolsep}{3.6pt}

\begin{tabular}{l c ccccccc}
\toprule
\textbf{Model}
& \textbf{Lang.}
& \textbf{Calculator}
& \textbf{Login Form}
& \textbf{Music Player}
& \textbf{Doc Editor}
& \textbf{Image Editor}
& \textbf{Info Card}
& \textbf{Overall} \\
\midrule

\multirow{2}{*}{Qwen3.5-4B$^{-}$}
& EN
& \cellcolor{qw4bg}2525.9
& \cellcolor{qw4bg}283.4
& \cellcolor{qw4bg}591.6
& \cellcolor{qw4bg}\underline{318.7}
& \cellcolor{qw4bg}\textbf{297.7}
& \cellcolor{qw4bg}\underline{192.9}
& \cellcolor{qw4overall}701.7 \\
& ZH
& \cellcolor{qw4bg}1252.4
& \cellcolor{qw4bg}\underline{198.3}
& \cellcolor{qw4bg}\underline{329.7}
& \cellcolor{qw4bg}\underline{285.2}
& \cellcolor{qw4bg}\underline{324.9}
& \cellcolor{qw4bg}\textbf{146.7}
& \cellcolor{qw4overall}\underline{422.9} \\

\multirow{2}{*}{Qwen3.5-4B$^{+}$}
& EN
& \cellcolor{qw4bg}9022.0
& \cellcolor{qw4bg}3625.2
& \cellcolor{qw4bg}4094.8
& \cellcolor{qw4bg}4570.7
& \cellcolor{qw4bg}5409.7
& \cellcolor{qw4bg}3232.2
& \cellcolor{qw4overall}4992.4 \\
& ZH
& \cellcolor{qw4bg}10181.8
& \cellcolor{qw4bg}4569.9
& \cellcolor{qw4bg}4607.1
& \cellcolor{qw4bg}5563.1
& \cellcolor{qw4bg}5612.4
& \cellcolor{qw4bg}4275.3
& \cellcolor{qw4overall}5801.6 \\
\midrule

\multirow{2}{*}{Qwen3.5-9B$^{-}$}
& EN
& \cellcolor{qw9bg}1991.9
& \cellcolor{qw9bg}309.6
& \cellcolor{qw9bg}393.1
& \cellcolor{qw9bg}340.1
& \cellcolor{qw9bg}298.0
& \cellcolor{qw9bg}\underline{192.9}
& \cellcolor{qw9overall}587.6 \\
& ZH
& \cellcolor{qw9bg}1427.6
& \cellcolor{qw9bg}299.6
& \cellcolor{qw9bg}426.1
& \cellcolor{qw9bg}376.6
& \cellcolor{qw9bg}447.4
& \cellcolor{qw9bg}\underline{203.5}
& \cellcolor{qw9overall}530.1 \\

\multirow{2}{*}{Qwen3.5-9B$^{+}$}
& EN
& \cellcolor{qw9bg}6866.7
& \cellcolor{qw9bg}\textbf{149.5}
& \cellcolor{qw9bg}\underline{292.9}
& \cellcolor{qw9bg}2826.7
& \cellcolor{qw9bg}3698.1
& \cellcolor{qw9bg}2773.8
& \cellcolor{qw9overall}2767.9 \\
& ZH
& \cellcolor{qw9bg}\textbf{596.5}
& \cellcolor{qw9bg}214.4
& \cellcolor{qw9bg}384.2
& \cellcolor{qw9bg}3578.6
& \cellcolor{qw9bg}4376.3
& \cellcolor{qw9bg}3645.1
& \cellcolor{qw9overall}2132.5 \\
\midrule

\multirow{2}{*}{Qwen3.5-27B$^{-}$}
& EN
& \cellcolor{qw27bg}\textbf{1009.7}
& \cellcolor{qw27bg}\underline{183.0}
& \cellcolor{qw27bg}\textbf{283.7}
& \cellcolor{qw27bg}\textbf{253.5}
& \cellcolor{qw27bg}\underline{297.8}
& \cellcolor{qw27bg}\textbf{136.4}
& \cellcolor{qw27overall}\textbf{360.7} \\
& ZH
& \cellcolor{qw27bg}\underline{961.1}
& \cellcolor{qw27bg}\textbf{188.4}
& \cellcolor{qw27bg}\textbf{304.0}
& \cellcolor{qw27bg}\textbf{264.2}
& \cellcolor{qw27bg}\textbf{323.7}
& \cellcolor{qw27bg}\textbf{146.7}
& \cellcolor{qw27overall}\textbf{364.7} \\

\multirow{2}{*}{Qwen3.5-27B$^{+}$}
& EN
& \cellcolor{qw27bg}6445.7
& \cellcolor{qw27bg}2340.8
& \cellcolor{qw27bg}3156.1
& \cellcolor{qw27bg}2796.6
& \cellcolor{qw27bg}3299.2
& \cellcolor{qw27bg}2555.2
& \cellcolor{qw27overall}3432.3 \\
& ZH
& \cellcolor{qw27bg}7029.6
& \cellcolor{qw27bg}3131.6
& \cellcolor{qw27bg}3536.1
& \cellcolor{qw27bg}3352.8
& \cellcolor{qw27bg}3947.7
& \cellcolor{qw27bg}3444.0
& \cellcolor{qw27overall}4073.6 \\
\midrule

\multirow{2}{*}{Qwen3-Coder-30B}
& EN
& \cellcolor{coderbg}\underline{1048.5}
& \cellcolor{coderbg}263.5
& \cellcolor{coderbg}404.9
& \cellcolor{coderbg}363.7
& \cellcolor{coderbg}420.7
& \cellcolor{coderbg}246.6
& \cellcolor{coderoverall}\underline{458.0} \\
& ZH
& \cellcolor{coderbg}1291.6
& \cellcolor{coderbg}355.5
& \cellcolor{coderbg}555.2
& \cellcolor{coderbg}496.1
& \cellcolor{coderbg}576.3
& \cellcolor{coderbg}220.1
& \cellcolor{coderoverall}582.5 \\
\midrule

\multirow{2}{*}{DeepSeek-R1}
& EN
& \cellcolor{dsbg}5651.8
& \cellcolor{dsbg}1792.8
& \cellcolor{dsbg}2192.3
& \cellcolor{dsbg}1304.4
& \cellcolor{dsbg}3023.1
& \cellcolor{dsbg}2009.2
& \cellcolor{dsoverall}2662.3 \\
& ZH
& \cellcolor{dsbg}5724.0
& \cellcolor{dsbg}2069.8
& \cellcolor{dsbg}1773.2
& \cellcolor{dsbg}3997.5
& \cellcolor{dsbg}3250.1
& \cellcolor{dsbg}1639.4
& \cellcolor{dsoverall}3075.7 \\
\bottomrule
\end{tabular}
\caption{Average raw response-token consumption (\(T_{\mathrm{out}}\)) across six GUI generation scenarios. EN and ZH denote the English and Chinese evaluation sets. Superscripts \(^{-}\) and \(^{+}\) indicate non-reasoning and reasoning-enabled configurations, respectively. Overall is the macro-average across the six scenarios. Within each language, the lowest and second-lowest values are shown in bold and underlined.}
\label{tab:response_token_consumption}
\end{table*}

\subsection{The Impact of Reasoning}
\label{sec:reasoning_impact}

\begin{figure}[H]
\centering
\includegraphics[width=\columnwidth]{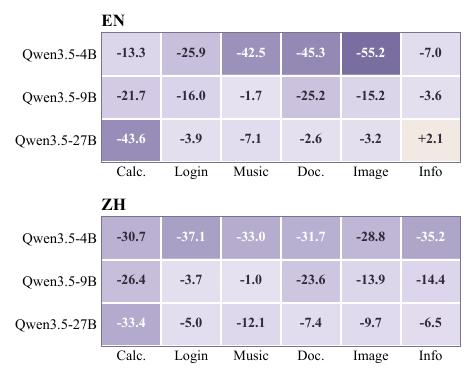}
\caption{Changes in GUI Score after enabling thinking mode for the Qwen3.5 models. Each value denotes \(\Delta=\mathrm{GS}_{\mathrm{thinking}}-\mathrm{GS}_{\mathrm{no\text{-}thinking}}\) in percentage points; negative values indicate performance degradation.}
\label{fig:thinking_mode_deltas}
\end{figure}

\textbf{Thinking mode generally reduces GUI generation performance, although its impact becomes smaller as model scale increases.} Figure~\ref{fig:thinking_mode_deltas} shows the change in GUI Score after enabling thinking mode for the three Qwen3.5 models across the six scenarios.

As shown in Figure~\ref{fig:thinking_mode_deltas}, Qwen3.5-4B exhibits the largest Overall decrease, dropping by 31.54 percentage points in EN and 32.74 percentage points in ZH. The corresponding reductions are 13.89 and 13.84 percentage points for Qwen3.5-9B, and 9.72 and 12.36 percentage points for Qwen3.5-27B. Thus, larger models are less affected by thinking mode, although none of the three models achieves an Overall improvement.

The impact also varies across layout structures. \textit{Calculator} declines substantially for all three models in both languages, while the larger variants remain relatively stable on structurally clearer scenarios such as \textit{Login Form} and \textit{Info Card}. The degradation is particularly widespread for Qwen3.5-4B, which also shows large reductions in \textit{Music Player}, \textit{Doc Editor}, and \textit{Image Editor}. The only positive change is the 2.12-point improvement of Qwen3.5-27B on the EN \textit{Info Card} scenario.

These results suggest that additional reasoning does not necessarily improve precise layout execution. It may instead introduce extra response content and increase the risk of deviating from the required output structure, particularly in dense-grid and multi-region layouts.

\subsection{Token Consumption in SchemaGUI}
\label{sec:response_token_consumption}

\textbf{Reasoning introduces substantial token overhead without consistent GUI Score gains, whereas larger models can achieve better performance with concise outputs.} We measure Response Tokens ($T_{\mathrm{out}}$) from the complete model response before parsing or JSON extraction, including reasoning content when present.

As shown in Table~\ref{tab:response_token_consumption}, structurally complex scenarios generally require more tokens. \textit{Calculator} has the highest average consumption at 3939.2 tokens, whereas \textit{Login Form} requires only 1248.5. Model scaling, however, does not necessarily increase response length: Qwen3.5-27B$^{-}$ consumes only 360.7 and 364.7 tokens in EN and ZH while achieving the highest Overall GUI Scores of 84.61\% and 81.78\%, respectively.

In contrast, Qwen3.5-4B$^{+}$ consumes 4992.4 tokens in EN and 5801.6 in ZH, compared with 701.7 and 422.9 for Qwen3.5-4B$^{-}$. Similar increases occur for the 9B and 27B models, indicating that reasoning overhead does not necessarily improve GUI Score. Detailed token ranges and clean-token statistics are provided in Appendix~\ref{sec:token_details}.

%% file: latex/06.Conclusion.01.tex
We introduce \textbf{SchemaGUI}, a template-based benchmark for controllable GUI generation evaluation. It produces paired natural-language instructions and deterministic function-call references across six GUI scenarios. Experiments show that larger models improve schema compliance and overall performance, but geometric layout control remains an important challenge. Geometry errors remain harder to resolve than text or structure mismatches, and explicit reasoning can hurt coordinate-dense layouts. These findings highlight the need for geometry-aware, schema-constrained evaluation.

%% file: latex/07.Limitation.02.tex
Despite its robustness, SchemaGUI has several limitations. First, it focuses on static structural synthesis and geometric layout planning, leaving dynamic user interactions and sequential runtime workflows unexplored. Second, the current evaluation follows a text-to-schema setting, while practical GUI design often depends on multimodal feedback loops that combine textual intent, visual inspection, and iterative refinement. Third, although the template-based design enables controllable and deterministic evaluation, it may not fully capture the visual diversity, stylistic variation, and open-ended constraints of real-world interfaces. In addition, our metrics mainly emphasize structural correctness and geometric fidelity, while subjective visual factors such as esthetics and design preference are not explicitly evaluated. Finally, although our experiments cover representative mainstream LLMs, future evaluations on more model families and larger sample sets may provide a more comprehensive understanding of model behavior. Future work will extend SchemaGUI toward interactive environments, richer GUI scenarios, human-centered visual assessment, and vision-language feedback for iterative layout self-correction.

%% file: latex/09_ethical_concern.tex
This work focuses on the construction and evaluation of a controlled benchmark for GUI generation. All samples in SchemaGUI are programmatically generated from predefined schemas and parameterized templates. The benchmark does not rely on real user interfaces, private screenshots, personal identifiers, user behavior logs, or any sensitive personal data. All annotations are automatically derived from deterministic templates, and the experiments are conducted in offline evaluation environments. Therefore, the current benchmark does not involve direct interaction with users or real-world decision-making processes.

The main ethical risks of this work are indirect. SchemaGUI is designed to evaluate structural correctness, geometric layout quality, and schema-following ability in GUI generation models. It should not be interpreted as a guarantee that a generated interface is safe, accessible, visually appropriate, or ready for deployment in user-facing systems. If similar GUI generation techniques are later used in practical design workflows, human review and additional validation would be necessary to prevent misleading layouts, inaccessible designs, or interfaces that fail to satisfy real user needs.

We also acknowledge that template-based generation provides strong controllability but may simplify the open-ended nature of real GUI design. As a result, the benchmark is intended as a diagnostic tool for measuring layout and structural generation ability, rather than a replacement for human-centered design evaluation. Future extensions involving real-world interfaces, interactive environments, or user-facing deployment should include stronger safeguards, such as privacy-aware data handling, accessibility checks, human evaluation, and careful review of generated interface behavior.

During the research process, AI tools may be used to support auxiliary tasks such as code debugging, writing refinement, and visualization preparation. All core research ideas, benchmark design choices, experimental analysis, and final conclusions are reviewed and determined by the authors. The use of AI assistance does not replace human responsibility for the correctness, validity, and ethical treatment of the research.

%% file: latex/08.Appendix.02.tex
\subsection{Scenario Schemas and Prompt Construction}
\label{sec:scenario_schemas}

SchemaGUI covers six GUI scenarios with different component structures and layout characteristics. \textit{Calculator} emphasizes compact grid organization; \textit{Login Form} and \textit{Info Card} primarily use sequential structures; \textit{Music Player} requires the coordinated organization of media information and playback controls; and \textit{Doc Editor} and \textit{Image Editor} contain multiple functional regions with more complex spatial relationships.

Within each scenario, diversity is introduced through variations in component composition, layout geometry, content attributes, and structural relations. The parameterized templates vary the window configuration, component inclusion, textual or data content, component dimensions, spacing, alignment, and scenario-specific organization while preserving the functional identity of the scenario. The EN and ZH subsets follow the same scenario schemas and structural controls, with their instructions and language-dependent content instantiated in the corresponding language. The resulting instances are sampled with fixed random seeds and deduplicated according to their deterministic reference layouts before evaluation.

Table~\ref{tab:config_diversity} summarizes the main sources of controlled variation within each scenario. These variations allow each scenario family to produce multiple interface configurations while preserving a consistent functional schema.

\begin{table}[H]
\centering
\small
\setlength{\tabcolsep}{4pt}
\begin{tabular}{lp{0.65\columnwidth}}
\toprule
Scenario & Controlled variation \\
\midrule
Calculator & Calculator type, grid structure, display configuration, operation set, and component scale. \\
Login Form & Form configuration, field composition, optional controls, layout style, window size, and spacing. \\
Music Player & Player layout, interface attributes, artwork, track metadata, playback controls, and feature set. \\
Doc Editor & Panel organization, directory structure, document content, toolbar actions, and interface attributes. \\
Image Editor & Canvas-panel layout, tool set, layer structure, selected layer, property fields, and interface attributes. \\
Info Card & Card layout, avatar, profile content, optional contact fields, window size, and spacing. \\
\bottomrule
\end{tabular}
\caption{Main sources of controlled variation across the six SchemaGUI scenarios.}
\label{tab:config_diversity}
\end{table}

Listing~\ref{lst:prompt_template} presents an abridged English representation of the common prompt structure used across the six scenarios. For each sampled design instance \(s\), the scenario specification, window and layout requirements, allowed functions, required components, and scenario-specific guidance are instantiated using the corresponding configuration. The ZH prompts follow the same structural requirements, with the instructions and language-dependent content expressed in Chinese.

\begin{center}
\begin{minipage}{\linewidth}
\begin{lstlisting}[
    basicstyle=\small\ttfamily,
    breaklines=true,
    frame=single,
    backgroundcolor=\color{gray!5},
    rulecolor=\color{gray!30},
    frameround=tttt,
    belowskip=1em,
    aboveskip=1em,
    captionpos=b,
    caption={Abridged English prompt illustrating the common structure used across the six GUI scenarios.},
    label={lst:prompt_template}
]
Return only a JSON array of function calls without explanations or additional text.

Scenario: {scenario}
Window and layout: {layout_specification}

Allowed functions:
{function_schemas}

Required components:
{component_requirements}

General layout requirements:
- Keep components within the specified window.
- Maintain consistent alignment and spacing.
- Separate different functional regions.

Scenario-specific requirements:
{scenario_guidance}

The first call must initialize the window.
Each subsequent call must conform to its function schema.
Return only the JSON array.
\end{lstlisting}
\end{minipage}
\end{center}

\paragraph{Prompt and reference construction.}
Each evaluation instance is derived from a shared design instance \(s\), which specifies the interface configuration, including its window, layout, component composition, content, and structural constraints. SchemaGUI uses the same \(s\) in two parallel branches. In the prompt branch, \(s\) is verbalized and inserted into the corresponding scenario template to produce the model input \(x\), from which the model generates the predicted function-call sequence \(y^L\). In the reference branch, \(s\) is deterministically expanded into the reference function-call sequence \(y^T\), which is hidden from the model and used only for evaluation.

This construction aligns the prompt \(x\) and deterministic reference \(y^T\) through their shared design instance while allowing the model to generate its own prediction \(y^L\) from the textual instruction. The prompt specifies the required components and layout conditions without exposing the complete reference function-call sequence. During evaluation, \(y^L\) is first checked for schema feasibility and is then compared with \(y^T\) using the GUI Score defined in Section~\ref{sec:schema_conditioned_eval}.

\begin{figure*}[t]
\centering
\includegraphics[width=\textwidth]{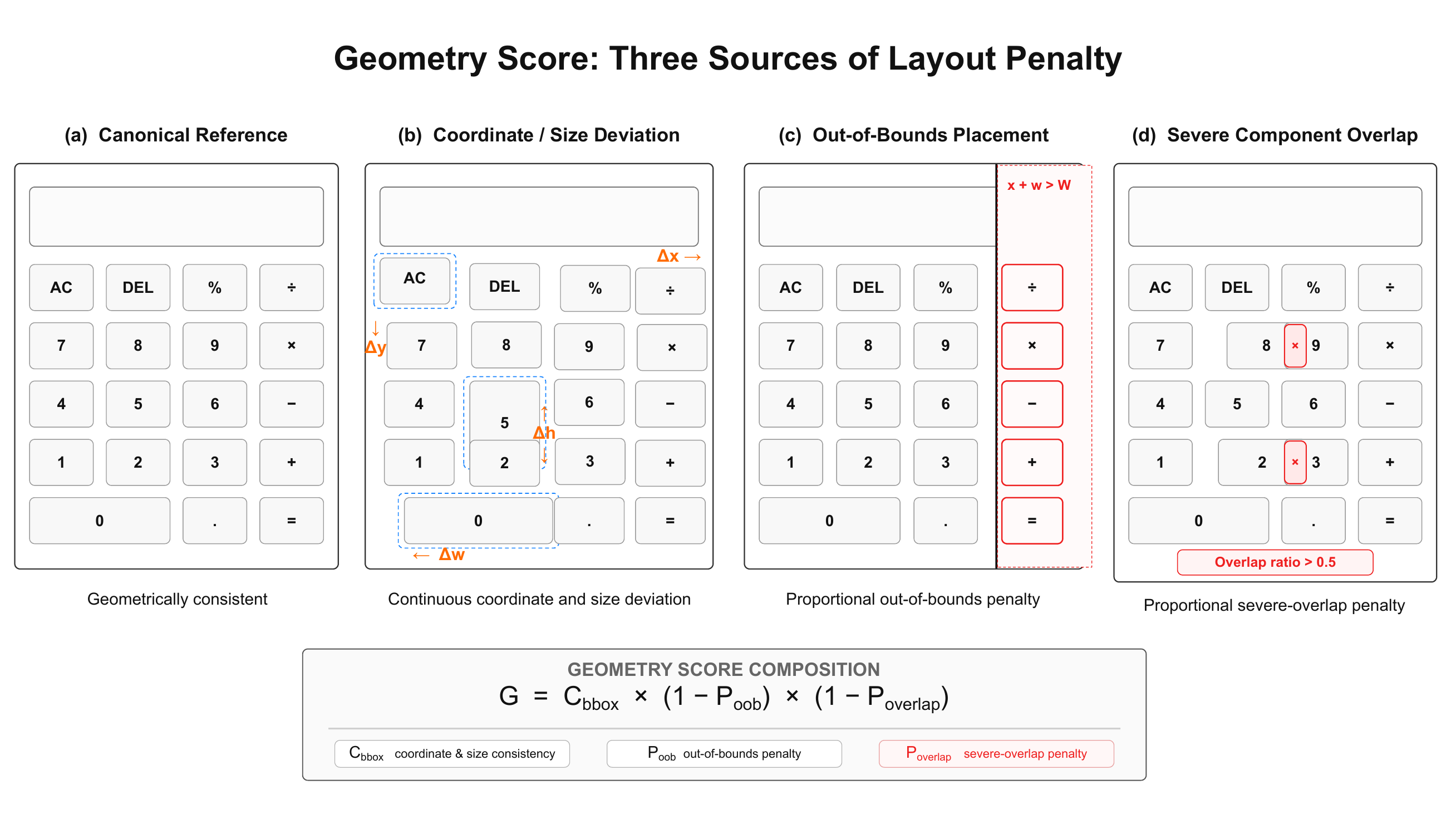}
\caption{Illustration of the Geometry score and its three penalty sources.}
\label{fig:geometry_score}
\end{figure*}

\subsection{Metric Implementation Details}
\label{app:metric_details}

This section provides further details on the automatic evaluation metrics introduced in Section~\ref{sec:schema_conditioned_eval}. For each instance, SchemaGUI generates a deterministic reference sequence \(y^T\) from the same schema configuration and slot values used to construct the model instruction.

\textbf{Schema feasibility.} The model prediction \(y^L\) is first parsed and examined for schema feasibility. Explanatory or reasoning content outside the generated function-call sequence is removed before evaluation, but only a complete and valid JSON array is accepted. A prediction is considered feasible only if it forms a non-empty function-call sequence, begins with a valid window-initialization call, uses supported component operations, and provides valid positions and positive component dimensions. Predictions that are empty, unparsable, incomplete, or inconsistent with the required schema are assigned a GUI Score of zero, with all five submetrics set to zero.

\textbf{Component matching.} Predicted components are aligned with the deterministic reference through a deterministic one-to-one matching procedure. Components are first matched by exact identity, defined by their component type and normalized textual content. Remaining components are matched only to reference components of the same type according to the Manhattan distance between their window-normalized center positions. Matching proceeds greedily from the closest candidate pair to the farthest. When multiple candidate pairs have the same distance, ties are resolved according to the original order of the reference and predicted components. This procedure prevents duplicate and cross-type matching and provides a shared correspondence for Geometry, Spatial Order, and Layout Harmony.

\textbf{Component F1.} Component F1 compares the multisets of component types in the predicted and reference sequences, excluding the window-initialization call. The multiset intersection determines the number of correctly recovered components, from which component-level precision and recall are calculated. Their harmonic mean gives the final Component F1 score, thereby penalizing both missing and redundant components.

\textbf{Text Match.} Text Match compares the multisets of component identities formed by each component type and its normalized \texttt{text} or \texttt{data} value. Text normalization removes capitalization differences, line-break variations, leading and trailing spaces, and redundant whitespace. The score is the proportion of reference textual identities recovered by the prediction, using exact normalized agreement rather than edit distance or semantic similarity.

\textbf{Spatial Order.} Spatial Order is computed from the visual arrangement of matched components rather than from their original function-call order. Components are grouped into the same visual row when the difference between their vertical center positions is no greater than either four pixels or half the height of the smaller component, whichever is larger. The rows are then ordered from top to bottom, and the components within each row are ordered from left to right. The score is the proportion of adjacent relations in the reference visual sequence that are preserved in the prediction. A relation receives no credit if either component is unmatched.

\textbf{Layout Harmony.} Layout Harmony \((\mathcal{H}_i)\) evaluates the internal visual organization of the generated interface from three complementary aspects: alignment, spacing, and proportion. The alignment term examines whether matched components preserve the reference relationships between their left, right, top, bottom, and center axes. Two axes are considered aligned when their difference is no greater than four pixels or one percent of the corresponding window dimension, whichever is larger. The spacing term compares the window-normalized horizontal and vertical gaps between neighboring components, with larger deviations receiving progressively lower scores. The proportion term evaluates whether neighboring components of the same type maintain relative width and height relationships consistent with the deterministic reference. The three terms are equally averaged:
\begin{equation}
\mathcal{H}_i =
\frac{
\mathcal{H}_i^{\mathrm{align}}
+
\mathcal{H}_i^{\mathrm{space}}
+
\mathcal{H}_i^{\mathrm{prop}}
}{3},
\end{equation}
where \(\mathcal{H}_i^{\mathrm{align}}\), \(\mathcal{H}_i^{\mathrm{space}}\), and \(\mathcal{H}_i^{\mathrm{prop}}\) denote the alignment, spacing, and proportion scores, respectively. Layout Harmony therefore complements Geometry by assessing the relationships among components rather than only the accuracy of their individual coordinates and sizes.

At the schema level, Layout Harmony partially captures visual quality through component alignment, spacing consistency, and proportional coherence. The bilingual EN and ZH instances evaluate language-dependent instructions and interface text, but do not explicitly model resource-based localization or locale-specific interface conventions. Rendering- and runtime-dependent properties, including font consistency, color contrast, accessibility, and security, are beyond the scope of the current offline evaluation.

\textbf{EN/ZH text handling.} The English (EN) and Chinese (ZH) settings determine the language of the instruction template. Textual slot values are preserved as instantiated and compared directly with the deterministic reference without translation. Because the model prediction and deterministic reference are derived from the same instance, Text Match evaluates language-aligned textual content under both settings.

Geometry is evaluated using the same component correspondence established above. It measures normalized coordinate and size consistency and further applies penalties for unmatched reference components, newly introduced out-of-bounds placement, and severe component overlap. Its complete formulation and penalty decomposition are provided in Appendix~\ref{app:geometry_score}.

\subsection{Geometry}
\label{app:geometry_score}

Section~\ref{sec:schema_conditioned_eval} introduces Geometry \((\mathcal{G}_i)\) as a continuous measure of coordinate and size consistency, together with penalties for out-of-bounds placement and severe component overlap. Here, \(i\) identifies an evaluation sample, \(T\) denotes the deterministic reference, and \(L\) denotes the model prediction. The Geometry score lies within \([0,1]\), with a higher value indicating better geometric consistency.

As illustrated in Figure~\ref{fig:geometry_score}, geometric evaluation is performed after the predicted components have been matched with the reference components. Let
\begin{equation}
\mathbf{b}_{ip}^{T}
=
\left(
x_{ip}^{T},
y_{ip}^{T},
w_{ip}^{T},
h_{ip}^{T}
\right)
\end{equation}
denote the bounding box of the \(p\)-th reference component in sample \(i\). Similarly, \(\mathbf{b}_{iq}^{L}\) denotes the bounding box of the \(q\)-th predicted component. The variables \(x\) and \(y\) represent the upper-left position of a component, while \(w\) and \(h\) represent its width and height.

For each valid matched pair \((p,q)\), the normalized bounding-box deviation is defined as
\begin{equation}
d_{i,pq}
=
\frac{
\left\|
\mathbf{b}_{ip}^{T}
-
\mathbf{b}_{iq}^{L}
\right\|_{1}
}{
W_i+H_i
},
\label{eq:bbox_deviation}
\end{equation}
where \(W_i\) and \(H_i\) are the width and height of the reference window in sample \(i\). The symbol \(\|\cdot\|_1\) denotes the sum of the absolute differences in \(x\), \(y\), \(w\), and \(h\). Thus, \(d_{i,pq}\) jointly measures position and size deviation relative to the scale of the complete interface.

Let \(n_i\) denote the number of reference components in sample \(i\). For each reference component \(p\), its geometric error is defined as
\begin{equation}
e_{ip}
=
\begin{cases}
d_{i,pq}, & \text{if \(p\) has a valid match \(q\)},\\
1, & \text{otherwise}.
\end{cases}
\label{eq:component_geometry_error}
\end{equation}
An unmatched reference component therefore contributes a unit error. The bounding-box consistency term is computed as
\begin{equation}
C_{\mathrm{bbox},i}
=
\max
\left\{
0,\,
1-
\frac{1}{n_i}
\sum_{p=1}^{n_i}
e_{ip}
\right\}.
\label{eq:bbox_consistency}
\end{equation}
Here, \(C_{\mathrm{bbox},i}\) denotes the bounding-box consistency of sample \(i\). The maximum operation ensures that the score does not fall below zero. Moderate coordinate or size deviations reduce the score continuously rather than causing the complete prediction to be rejected.

The bounding-box consistency term is further adjusted by two penalties. The out-of-bounds penalty \(P_{\mathrm{oob},i}\) is the proportion of predicted components that introduce placement outside the reference window. A predicted component contributes to this penalty when it extends beyond the window while its matched reference component does not. An unmatched predicted component is also penalized if it extends beyond the window. When no predicted component has a valid bounding box, \(P_{\mathrm{oob},i}\) is set to \(1\).

The overlap penalty \(P_{\mathrm{overlap},i}\) is the proportion of same-type predicted component pairs that introduce severe overlap not present between their corresponding reference components. When no same-type predicted component pair is available, \(P_{\mathrm{overlap},i}\) is set to \(0\).

The final Geometry score is computed as
\begin{equation}
\mathcal{G}_i
=
C_{\mathrm{bbox},i}
\left(
1-P_{\mathrm{oob},i}
\right)
\left(
1-P_{\mathrm{overlap},i}
\right),
\label{eq:geometry_decomposition}
\end{equation}
where \(\mathcal{G}_i\) is the final Geometry score for sample \(i\), \(C_{\mathrm{bbox},i}\) measures bounding-box consistency, \(P_{\mathrm{oob},i}\) measures newly introduced out-of-bounds placement, and \(P_{\mathrm{overlap},i}\) measures newly introduced severe overlap.

To define severe overlap, let \(\mathbf{b}_{ir}^{L}\) and \(\mathbf{b}_{is}^{L}\) denote the bounding boxes of two distinct predicted components, indexed by \(r\) and \(s\), in sample \(i\). Their overlap ratio is
\begin{equation}
\operatorname{OR}
\left(
\mathbf{b}_{ir}^{L},
\mathbf{b}_{is}^{L}
\right)
=
\frac{
\operatorname{Area}
\left(
\mathbf{b}_{ir}^{L}
\cap
\mathbf{b}_{is}^{L}
\right)
}{
\min
\left\{
\operatorname{Area}
\left(
\mathbf{b}_{ir}^{L}
\right),
\operatorname{Area}
\left(
\mathbf{b}_{is}^{L}
\right)
\right\}
},
\label{eq:overlap_ratio}
\end{equation}
where \(\operatorname{OR}(\cdot,\cdot)\) denotes the overlap ratio, \(\operatorname{Area}(\cdot)\) denotes the area of a component box, and \(\cap\) denotes the intersection of two boxes. An overlap is considered severe when
\begin{equation}
\operatorname{OR}
\left(
\mathbf{b}_{ir}^{L},
\mathbf{b}_{is}^{L}
\right)
>
0.5.
\end{equation}
This condition means that more than half of the smaller component is covered. Only newly introduced severe overlap is penalized, which prevents a prediction from being penalized for an overlap relationship already present in the deterministic reference.

This decomposition preserves the continuous geometric evaluation described in Section~\ref{sec:schema_conditioned_eval}. Isolated coordinate or size differences produce moderate reductions in \(\mathcal{G}_i\), whereas widespread boundary violations or severe component overlap lead to larger reductions.

\begin{table}[t]
\centering
\small
\setlength{\tabcolsep}{3.2pt}
\begin{tabular*}{\columnwidth}{@{\extracolsep{\fill}}lcccc}
\toprule
Scenario & \(\mathcal{G}\) at \(0.5\) & Min. \(\mathcal{G}\) & Max. \(\mathcal{G}\) & Range \\
\midrule
Calculator   & 0.51070 & 0.51060 & 0.51074 & 0.00014 \\
Login Form   & 0.82953 & 0.82947 & 0.82953 & 0.00007 \\
Music Player & 0.67393 & 0.67304 & 0.67398 & 0.00094 \\
Doc Editor   & 0.81896 & 0.81704 & 0.81896 & 0.00191 \\
Image Editor & 0.66628 & 0.66598 & 0.66661 & 0.00063 \\
Info Card    & 0.67729 & 0.67620 & 0.67730 & 0.00110 \\
\midrule
Overall      & 0.69612 & 0.69539 & 0.69619 & 0.00080 \\
\bottomrule
\end{tabular*}
\caption{Sensitivity of Geometry to the severe-overlap threshold. The threshold is varied from \(0.1\) to \(0.9\) in increments of \(0.1\). The symbol \(\mathcal{G}\) denotes the average Geometry score. Results are averaged over the five evaluated models in both EN and ZH settings, with \(1{,}000\) samples per model, language, and scenario.}
\label{tab:overlap_threshold_sensitivity}
\end{table}

\paragraph{Sensitivity to the severe-overlap threshold.}
We examine the sensitivity of Geometry to the threshold used to identify severe component overlap. Table~\ref{tab:overlap_threshold_sensitivity} reports the average Geometry score \(\mathcal{G}\) at the default threshold of \(0.5\), together with the minimum, maximum, and range obtained when the threshold is varied from \(0.1\) to \(0.9\). Geometry remains stable across all six scenarios. The largest variation is \(0.00191\) for Doc Editor, while the overall variation is only \(0.00080\). This limited sensitivity indicates that few component pairs have overlap ratios close to the decision boundary. Most geometric differences are instead captured by continuous coordinate and size deviations, unmatched reference components, and out-of-bounds placement. We therefore use \(0.5\) as the default threshold, under which an overlap is considered severe when more than half of the smaller component is covered.
\subsection{Error Analysis on the Calculator Scenario}
\label{app:calculator_error_analysis}

\begin{figure*}[t]
\centering
\includegraphics[width=\textwidth]{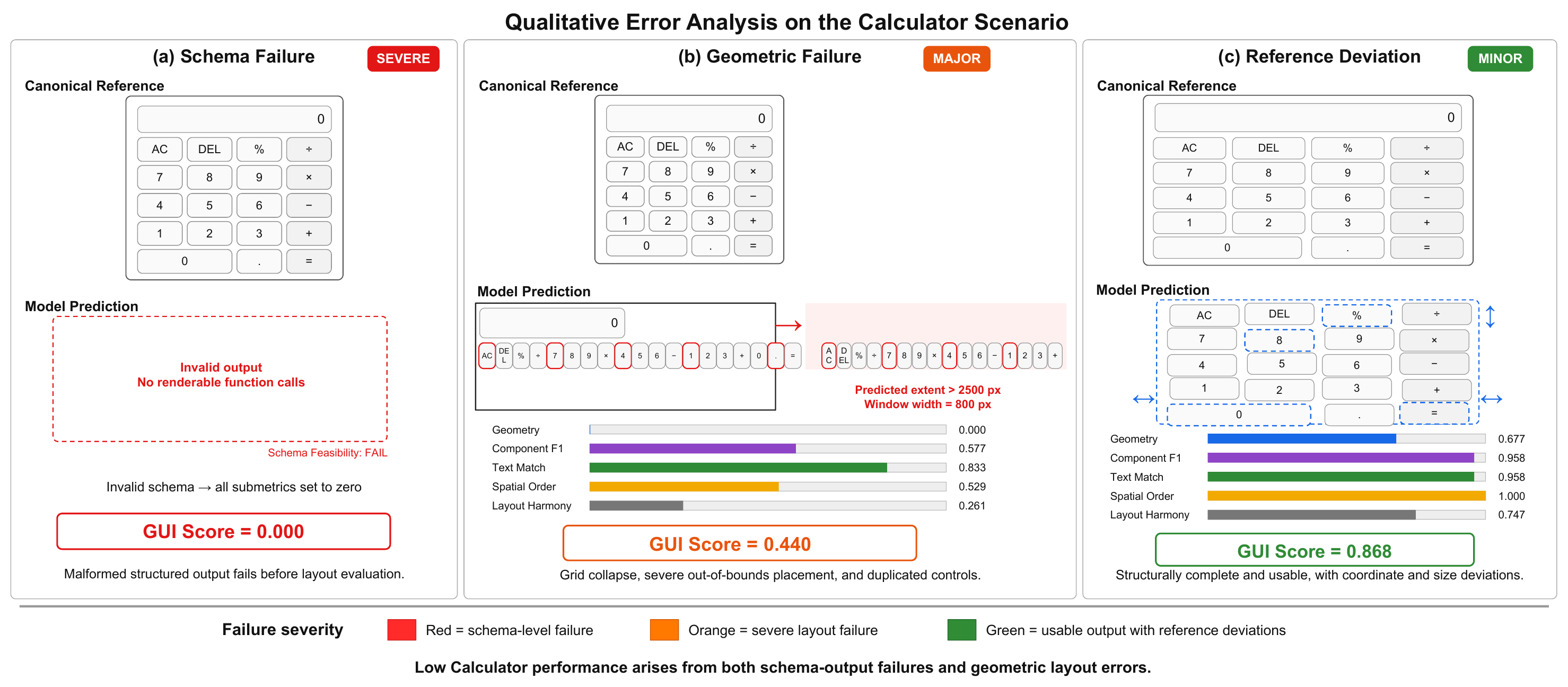}
\caption{Representative error cases in the Calculator scenario. (a) A schema-infeasible output receives a GUI Score of zero. (b) A schema-feasible prediction suffers from grid collapse, severe out-of-bounds placement, and duplicated controls. (c) A structurally complete and usable prediction retains a relatively high GUI Score despite coordinate and size deviations from the deterministic reference.}
\label{fig:calculator_error_analysis}
\end{figure*}

The Calculator scenario provides a useful setting for examining generation errors because it requires many visually similar controls to be organized into a dense and regular two-dimensional grid. Figure~\ref{fig:calculator_error_analysis} summarizes three representative error categories observed in this setting: schema-level failure, geometric layout failure, and reference deviation.

\paragraph{Schema-level failure.}
Schema-level failure occurs when the generated response cannot be converted into a valid function-call sequence, for example because of malformed structured output or missing required arguments. Such an output fails the Schema Feasibility check before reference-based evaluation and is assigned zero across all submetrics, following the evaluation hierarchy defined in Section~\ref{sec:schema_conditioned_eval}. This category primarily reflects structured-output instability and instruction-following failure rather than geometric inaccuracy.

\paragraph{Geometric layout failure.}
Passing schema validation does not guarantee a usable interface. A schema-valid prediction may preserve recognizable component types and text while exhibiting grid collapse, extensive out-of-bounds placement, severe overlap, or highly inconsistent dimensions. Other errors, such as duplicated controls, may occur simultaneously and further reduce component-level fidelity.

As illustrated in Figure~\ref{fig:calculator_error_analysis}(b), the two-dimensional calculator grid can collapse into an elongated sequence that extends far beyond the visible window. In this condition, Component F1, Text Match, Spatial Order, and Layout Harmony may retain partial credit because some intended content and structural relationships remain recognizable, while Geometry is strongly reduced by the invalid spatial arrangement. This demonstrates why schema validity and component recovery alone are insufficient for measuring GUI usability.

\paragraph{Reference deviation.}
Reference deviation represents an intermediate condition in which the generated interface remains structurally complete and practically usable but differs from the deterministic reference in coordinates, dimensions, spacing, or grid proportions. These differences reduce \(C_{\mathrm{bbox}}\) without invalidating the complete output.

Figure~\ref{fig:calculator_error_analysis}(c) illustrates this distinction. Most expected controls and their structural relationships are retained, but their placement and dimensions do not precisely follow the reference. Compared with a global layout failure, this type of output receives a more moderate Geometry penalty and preserves a relatively high GUI Score.

These categories clarify the complementary roles of the benchmark submetrics. Schema Feasibility determines whether an output is executable; Component F1 and Text Match measure whether the expected controls and content are recovered; Spatial Order and Layout Harmony evaluate their relative organization; and Geometry evaluates whether matched controls are placed and sized consistently. The evaluation can therefore distinguish outputs that cannot be rendered, outputs that are executable but spatially unusable, and outputs that remain usable while deviating from the deterministic reference.

The Calculator scenario makes these distinctions particularly visible. Recovering the button labels is not sufficient: successful generation also requires consistent row--column alignment, component dimensions, spacing, and window containment.

\subsection{Detailed Token Statistics}
\label{sec:token_details}

Table~\ref{tab:raw_token_extrema} and Table~\ref{tab:clean_token_consumption} provide complementary views of response-side token usage. Since all models are evaluated with the same scenario-specific prompts and slot descriptions, input token counts are largely determined by the benchmark setting. We therefore focus on response tokens, which better reflect model behavior, generation stability, and the cost of valid GUI outputs.

\begin{table*}[t]
\centering
\small
\renewcommand{\arraystretch}{1.15}
\setlength{\tabcolsep}{5.0pt}
\resizebox{\textwidth}{!}{
\begin{tabular}{l c cccccc}
\toprule
\textbf{Model}
& \textbf{Lang.}
& \textbf{Calculator}
& \textbf{Login Form}
& \textbf{Music Player}
& \textbf{Doc Editor}
& \textbf{Image Editor}
& \textbf{Info Card} \\
\midrule

\multirow{2}{*}{Qwen3.5-4B$^{-}$}
& EN & 692--6778 & 18--3406 & 18--3574 & 192--500 & 291--325 & 149--238 \\
& ZH & 112--5741 & 141--400 & 239--3649 & 196--521 & 297--515 & 109--189 \\

\multirow{2}{*}{Qwen3.5-4B$^{+}$}
& EN & 5433--11715 & 0--7832 & 2570--7713 & 2302--7815 & 3556--7686 & 1161--7812 \\
& ZH & 5242--12480 & 2438--8735 & 3172--8521 & 3741--8418 & 3615--8194 & 2334--8951 \\
\midrule

\multirow{2}{*}{Qwen3.5-9B$^{-}$}
& EN & 664--4463 & 162--525 & 225--3453 & 192--620 & 291--326 & 149--238 \\
& ZH & 650--7199 & 199--579 & 335--515 & 278--540 & 434--596 & 153--259 \\

\multirow{2}{*}{Qwen3.5-9B$^{+}$}
& EN & 3367--7913 & 0--526 & 0--490 & 0--7661 & 2116--7755 & 1194--7968 \\
& ZH & 0--2397 & 0--378 & 0--515 & 0--8270 & 2869--8698 & 1951--8857 \\
\midrule

\multirow{2}{*}{Qwen3.5-27B$^{-}$}
& EN & 578--6811 & 137--258 & 225--342 & 192--331 & 291--324 & 105--169 \\
& ZH & 578--6781 & 141--269 & 239--367 & 196--381 & 312--356 & 109--189 \\

\multirow{2}{*}{Qwen3.5-27B$^{+}$}
& EN & 3521--7680 & 1303--7618 & 1788--6175 & 1784--3711 & 2408--6712 & 1211--4644 \\
& ZH & 3453--7707 & 1952--3833 & 2416--7193 & 2283--3926 & 3000--7984 & 1609--7867 \\
\midrule

\multirow{2}{*}{Qwen3-Coder}
& EN & 224--1929 & 194--569 & 310--643 & 274--477 & 293--600 & 137--315 \\
& ZH & 880--3238 & 262--755 & 438--670 & 367--707 & 562--633 & 154--336 \\
\midrule

\multirow{2}{*}{DeepSeek-R1}
& EN & 0--9642 & 0--5120 & 0--6055 & 0--6075 & 0--7905 & 0--8651 \\
& ZH & 0--10978 & 0--6003 & 0--6812 & 0--8465 & 608--7308 & 345--6926 \\
\bottomrule
\end{tabular}
}
\caption{Observed minimum--maximum ranges of raw response-token counts. This table includes all generations, including failed, empty, and timed-out outputs. A minimum of zero indicates that at least one sample produced no recorded raw response tokens. Superscripts $^{-}$ and $^{+}$ denote non-reasoning and reasoning-enabled configurations, respectively.}
\label{tab:raw_token_extrema}
\end{table*}

\begin{table*}[t]
\centering
\small
\renewcommand{\arraystretch}{1.15}
\setlength{\tabcolsep}{4.5pt}
\resizebox{\textwidth}{!}{
\begin{tabular}{l c ccccccc}
\toprule
\textbf{Model}
& \textbf{Lang.}
& \textbf{Calculator}
& \textbf{Login Form}
& \textbf{Music Player}
& \textbf{Doc Editor}
& \textbf{Image Editor}
& \textbf{Info Card}
& \textbf{Overall} \\
\midrule

\multirow{2}{*}{Qwen3.5-4B$^{-}$}
& EN & 1278.6 & 182.3 & 285.3 & 254.5 & 297.7 & 192.9 & 415.2 \\
& ZH & 873.8 & 194.7 & 309.0 & 264.5 & 324.9 & 146.7 & 352.3 \\

\multirow{2}{*}{Qwen3.5-4B$^{+}$}
& EN & 193.4 & 154.2 & 175.6 & 220.0 & 183.6 & 178.8 & 184.2 \\
& ZH & 1293.6 & 173.3 & 282.2 & 322.8 & 369.7 & 145.4 & 431.1 \\
\midrule

\multirow{2}{*}{Qwen3.5-9B$^{-}$}
& EN & 1118.6 & 185.6 & 291.1 & 256.4 & 298.0 & 192.9 & 390.4 \\
& ZH & 1241.3 & 278.4 & 426.1 & 376.6 & 447.4 & 203.5 & 495.5 \\

\multirow{2}{*}{Qwen3.5-9B$^{+}$}
& EN & 491.9 & 189.0 & 301.9 & 236.6 & 230.5 & 179.4 & 271.6 \\
& ZH & 1218.1 & 228.1 & 388.0 & 287.0 & 355.5 & 168.1 & 440.8 \\
\midrule

\multirow{2}{*}{Qwen3.5-27B$^{-}$}
& EN & 871.8 & 183.0 & 283.7 & 253.5 & 297.8 & 136.4 & 337.7 \\
& ZH & 899.2 & 188.4 & 304.0 & 264.2 & 323.7 & 146.7 & 354.3 \\

\multirow{2}{*}{Qwen3.5-27B$^{+}$}
& EN & 419.1 & 186.8 & 287.7 & 274.0 & 325.5 & 177.5 & 278.4 \\
& ZH & 897.8 & 180.6 & 270.4 & 252.4 & 299.0 & 172.9 & 345.5 \\
\midrule

\multirow{2}{*}{Qwen3-Coder}
& EN & 1048.5 & 261.2 & 404.9 & 363.7 & 420.7 & 246.6 & 457.6 \\
& ZH & 1282.4 & 351.7 & 555.2 & 496.1 & 576.3 & 220.1 & 580.3 \\
\midrule

\multirow{2}{*}{DeepSeek-R1}
& EN & 1279.2 & 184.2 & 286.7 & 260.8 & 310.0 & 177.1 & 416.3 \\
& ZH & 1339.5 & 189.5 & 304.4 & 283.6 & 338.4 & 194.6 & 441.7 \\
\bottomrule
\end{tabular}
}
\caption{Average clean JSON-token counts for successfully parsed GUI outputs. Clean-token statistics include only valid generations that can be parsed into GUI specifications, while parse-failed, empty, and timed-out outputs are excluded. Overall is the macro-average across the six scenarios.}
\label{tab:clean_token_consumption}
\end{table*}

The two token tables capture different aspects of generation behavior. Table~\ref{tab:raw_token_extrema} reports raw response-token ranges over all generations, including failed and empty outputs. It therefore reflects the instability of model responses under strict structured-output constraints. Large maximum values, especially in reasoning-enabled configurations, indicate that models may spend many tokens on intermediate reasoning or malformed responses before producing a valid GUI specification. Minimum values of zero correspond to cases where no usable response was recorded, such as empty outputs or timeouts.

In contrast, Table~\ref{tab:clean_token_consumption} focuses only on successfully parsed outputs. It measures the token cost of valid GUI specifications after cleaning and parsing, and is therefore more suitable for estimating the cost of usable generations. The clean-token statistics show that calculator often requires the longest valid outputs, while login form and info card are generally more compact. This suggests that response length is strongly affected by the structural complexity and constraint density of each GUI scenario.

Comparing the two tables further reveals the gap between raw generation cost and usable output cost. Reasoning-enabled models may exhibit much larger raw-token ranges, but this does not necessarily translate into longer valid JSON outputs or better GUI quality. This supports our observation that structured GUI generation benefits more from concise format adherence than from unconstrained long-form reasoning. Therefore, we report both raw-token extrema and clean-token consumption: the former diagnoses unstable generation behavior, while the latter reflects the actual token cost of valid GUI outputs.

\begin{table}[H]
\centering
\small
\setlength{\tabcolsep}{3pt}
\resizebox{\columnwidth}{!}{
\begin{tabular}{lccccc}
\toprule
Model & Render & Fidelity & Layout & Visual & Human \\
\midrule
Qwen3.5-4B & 72.50 & 3.86 & 3.38 & 2.91 & 49.06 \\
Qwen3.5-9B & 88.30 & \underline{4.02} & 3.34 & 2.68 & 59.10 \\
Qwen3.5-27B & \underline{94.19} & 4.16 & \textbf{3.93} & \textbf{3.42} & \underline{72.28} \\
Qwen3-Coder & \textbf{100.00} & \textbf{4.41} & \underline{3.69} & \underline{3.18} & \textbf{75.20} \\
DeepSeek-R1 & 90.80 & 3.70 & 3.33 & 2.91 & 60.18 \\
\bottomrule
\end{tabular}
}
\caption{Human evaluation results across five models. The best and second-best results are highlighted in bold and underlined, respectively.}
\label{tab:human_eval}
\end{table}

\subsection{Human Evaluation}

To validate whether the automatic metrics align with human perception of GUI quality, we conduct a small-scale blind human evaluation of 1,200 generated GUIs, covering five models, six GUI scenarios, and two languages. For each model-scenario-language combination, we randomly sample 20 generated results and show only the rendered GUI images to the evaluator, without revealing model identities or automatic scores.

Three evaluators participated in the study. The 1,200 rendered GUIs were randomly and evenly assigned across the evaluators, with each evaluator independently assessing 400 samples. The assignments were balanced across models, scenarios, and languages to reduce evaluator-specific bias. All evaluators received reasonable compensation for their participation.

Each GUI is rated on a 1-5 scale from three aspects: requirement fidelity, layout usability, and visual quality. The final human score for each GUI is computed as the average of these three ratings and then normalized to a 0-100 scale. Failed or non-renderable generations are assigned a score of zero.

As shown in Table~\ref{tab:human_eval}, the human evaluation results are broadly consistent with the automatic GUI scores. Models with stronger automatic performance generally receive higher human ratings. In particular, Qwen3-Coder and Qwen3.5-27B obtain the top two human scores, which is consistent with their leading performance in the automatic evaluation. In contrast, Qwen3.5-4B receives the lowest human score, reflecting its lower rendering success rate and weaker perceived GUI quality.

The comparison also shows that human judgments and automatic metrics are not identical but complementary. Fidelity scores are generally higher than layout and visual scores, suggesting that models can often include the required UI components while still struggling with spatial arrangement and visual polish. This trend is consistent with the automatic evaluation, where geometry-related errors remain a major source of quality degradation.

To further quantify the agreement between automatic and human evaluation, we compute Kendall's $\tau$ and Spearman's $\rho$, two standard rank correlation measures, between the automatic overall GUI score and the human score across the five evaluated models. The resulting Kendall's $\tau=0.80$ and Spearman's $\rho=0.90$ indicate a high rank-level agreement between automatic and human evaluation. These results support the reliability of our automatic evaluation protocol for scalable GUI generation benchmarking.

\subsection{Bilingual Analysis}
\label{sec:bilingual_analysis}

Table~\ref{tab:main_gui_results} shows that the effect of language is model- and scenario-dependent. On average, Qwen3.5-4B, Qwen3.5-9B, and DeepSeek-R1 perform better on the Chinese set, with overall gains of 3.11, 2.38, and 4.33 points. In contrast, Qwen3.5-27B and Qwen3-Coder-30B obtain slightly higher overall scores on the English set, with Chinese scores lower by 2.83 and 2.24 points. Thus, the bilingual gap is not simply explained by model size.

The clearest language difference appears in calculator. All five models improve on the Chinese set, especially Qwen3.5-4B and Qwen3.5-9B, whose scores increase from 33.25 to 56.26 and from 43.40 to 63.22. Login form shows a similar but weaker pattern, with all models matching or slightly improving in Chinese. In contrast, image editor favors English for most models: Qwen3.5-4B, Qwen3.5-27B, Qwen3-Coder-30B, and DeepSeek-R1 all drop by 7--12 points on the Chinese set. Doc editor also mostly favors English, with DeepSeek-R1 being a notable exception, improving from 39.41 to 75.00 in Chinese.

The token statistics provide a useful check on these trends. Chinese outputs often contain more clean JSON tokens, especially for Qwen3.5-9B and Qwen3-Coder-30B. For example, Qwen3-Coder-30B increases from 457.6 average clean tokens in English to 580.3 in Chinese, while Qwen3.5-9B$^{-}$ increases from 390.4 to 495.5. This extra length, however, does not always translate into better GUI quality: Qwen3-Coder-30B uses more clean tokens in Chinese but obtains a lower overall GUI score. This suggests that longer valid JSON outputs may reflect more verbose specifications rather than better layouts.

Raw token ranges also show that reasoning-enabled settings can be less stable across languages. Several reasoning-enabled models have much wider raw-token ranges, including zero-token cases and very long responses, indicating empty outputs, timeouts, or excessive intermediate generation. These effects are most visible in calculator and doc editor, which are also among the more language-sensitive scenarios.

Overall, the bilingual results suggest that language affects both GUI quality and generation efficiency. Chinese prompts help in compact, constraint-driven scenarios such as calculator and login form, while English prompts are generally more stable in visually complex scenarios such as image editor.